\documentclass[runningheads]{llncs}

\usepackage{accv}

\usepackage{accvabbrv}
\usepackage{makecell}
\usepackage{graphicx}
\usepackage{booktabs}
\usepackage{amssymb}
\usepackage{multirow}
\usepackage{wrapfig}
\usepackage[accsupp]{axessibility}  
\newcommand{\Tref}[1]{Table~\ref{#1}}

\newcommand{\fref}[1]{Fig.~\ref{#1}}
\newcommand{\Fref}[1]{Figure~\ref{#1}}

\newcommand{\textblock}[1]{\noindent\textbf{#1}}

\usepackage[pagebackref,breaklinks,colorlinks,citecolor=accvblue]{hyperref}

\usepackage{orcidlink}

\begin{document}

\title{ATTIQA: Generalizable Image Quality Feature Extractor using Attribute-aware Pretraining}

\titlerunning{ATTIQA}

\author{Daekyu Kwon\inst{1}\orcidlink{0009-0006-2154-3348} \and
Dongyoung Kim\inst{1}\orcidlink{0009-0000-6414-2380} \and
Sehwan Ki\inst{2}\orcidlink{0000-0002-3809-7886} \and
Younghyun Jo\inst{2}\orcidlink{0000-0002-8530-9802} \and \\
Hyong-Euk Lee\inst{2} \and
Seon Joo Kim\inst{1}\orcidlink{0000-0001-8512-216X}}

\authorrunning{Kwon et al.}

\institute{Yonsei University \and
Samsung Advanced Institue of Technology}

\maketitle

\begin{abstract}
In no-reference image quality assessment (NR-IQA), the challenge of limited dataset sizes hampers the development of robust and generalizable models. 
Conventional methods address this issue by utilizing large datasets to extract rich representations for IQA.
Also, some approaches propose vision language models (VLM) based IQA, but the domain gap between generic VLM and IQA constrains their scalability.
In this work, we propose a novel pretraining framework that constructs a generalizable representation for IQA by selectively extracting quality-related knowledge from VLM and leveraging the scalability of large datasets.
Specifically, we select optimal text prompts for five representative image quality attributes and use VLM to generate pseudo-labels.
Numerous attribute-aware pseudo-labels can be generated with large image datasets, allowing our IQA model to learn rich representations about image quality.
Our approach achieves state-of-the-art performance on multiple IQA datasets and exhibits remarkable generalization capabilities.
Leveraging these strengths, we propose several applications, such as evaluating image generation models and training image enhancement models, demonstrating our model's real-world applicability.
\end{abstract}
\newcommand{\textred}[1]{\textcolor{red}{#1}}

\section{Introduction}
\label{sec:intro}
No-reference image quality assessment (NR-IQA) \cite{mittal2012no, talebi2018nima, ye2012unsupervised, zhu2020metaiqa, golestaneh2022no, su2020blindly} is a task of quantifying the quality of images without a pristine reference image.
Recently, methods in IQA have also started incorporating deep learning  \cite{zhang2018unreasonable,bosse2017deep,liu2017rankiqa,zhang2018blind,kang2014convolutional,you2021transformer}, similar to other fields in computer vision.
However, effective application of deep learning in image quality assessment (IQA) faces challenges due to the limited size of existing IQA datasets \cite{ponomarenko2015image, ying2020patches, hosu2020koniq, fang2020perceptual, ghadiyaram2015massive}.
Training an IQA model from scratch with a small dataset encounters difficulties in learning rich representations for image quality.
This often results in degraded performance and poor generalization, thereby restricting the practical applicability of IQA in real-world scenarios.

\begin{figure}[t]
\centering
\includegraphics[clip, trim={0 0 0 0}, width=1.0\textwidth]{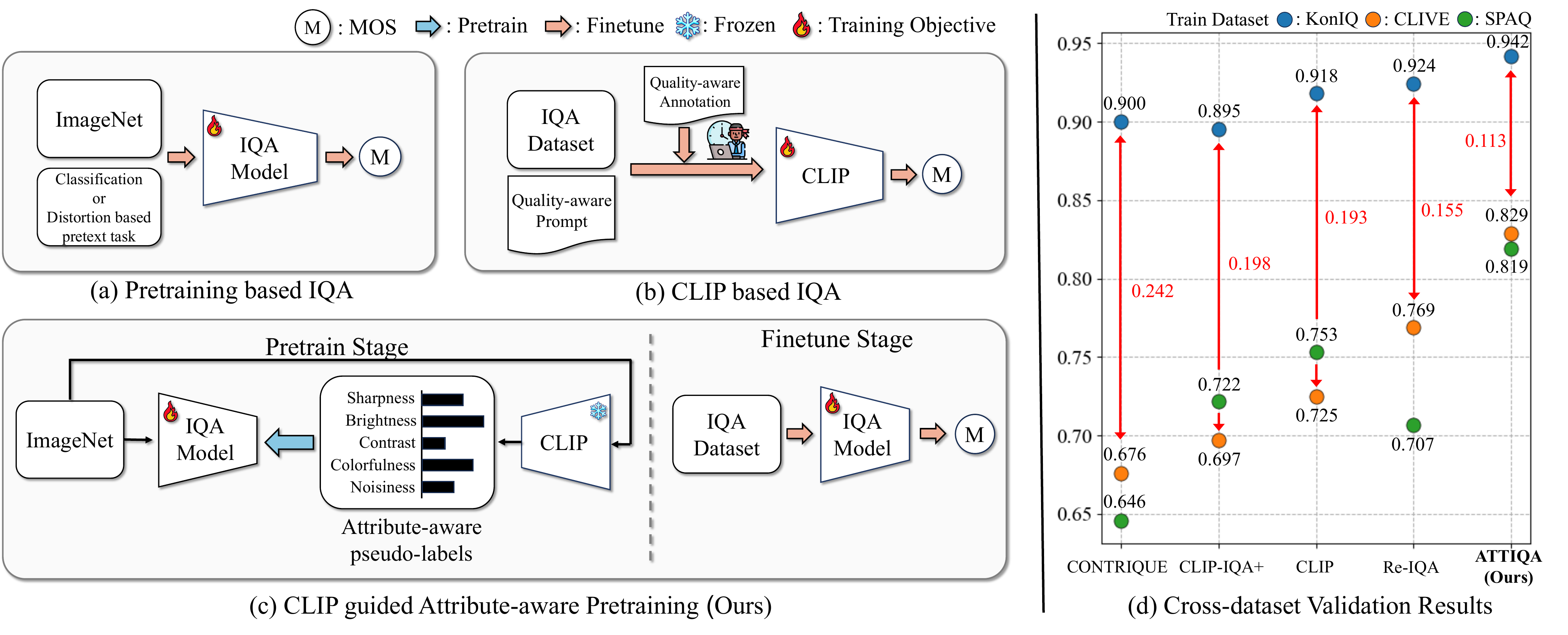}
\caption{An illustration of a training strategy for IQA across previous works and ours.
(a) Classic IQA models use ImageNet-pretrained models or suggest image quality-related pretraining.
(b) CLIP-based IQA directly utilizes CLIP or adapts it for IQA using additional quality annotations, which requires human labor.
(c) Our method incorporates the rich representation of large datasets and leverages CLIP's IQA capability.
We pretrain IQA model with attribute-aware pseudo-labels derived from CLIP and finetune it to the target IQA dataset.
(d) Cross dataset validation results, obtained by testing on the KonIQ dataset after training on various datasets.
ATTIQA achieves state-of-the-art results and exhibits superior generalization capability on unseen datasets, showing less performance decline on cross-dataset setup compared to other methods.
}
\label{fig:teaser}
\end{figure}

To address the generalization issue derived from limited dataset size, IQA approaches have been developed to leverage rich representations from large datasets~\cite{russakovsky2014imagenet} (\fref{fig:teaser}(a)).
In~\cite{9271914,talebi2018nima, zhang2018blind, bianco2018use}, transfer learning strategy was utilized by pretraining a model on ImageNet~\cite{russakovsky2014imagenet}. 
Several studies~\cite{zhang2018blind, zhao2023quality, saha2023re, liu2017rankiqa, madhusudana2022image} have proposed IQA-specific pretext tasks, founded on the premise that distortions in images directly impact their quality.
These lines of research emphasize the importance of pretraining tasks in IQA, demonstrating the benefits of the scalability of large datasets.
However, research on how to efficiently extract quality-related representations from large datasets is still in progress.

Recently, Vision Language Models (VLM), exemplified by CLIP~\cite{radford2021learning}, have emerged as a robust backbone in computer vision, highlighting their generalization capabilities.
Building on these strengths, exploiting VLM for IQA has also been explored (\fref{fig:teaser}(b)).
CLIP-IQA~\cite{wang2023exploring} proposed zero-shot IQA using the quality-aware prompt, showing the applicability of CLIP for IQA.
While CLIP-based IQA demonstrates good generalization capability without fine-tuning, it has been noted that CLIP alone is not suitable for precise IQA tasks, as it is trained on generic image-text pairs.
To address this issue, several studies attempt to adapt CLIP to the IQA domain using text prompts related to image quality.
Although these approaches effectively enhance CLIP's representation for IQA, these strategies are constrained by the necessity of supplementary image-text pairs for direct CLIP training, which requires additional human labor.

In this work, we introduce a novel pretraining framework for IQA, named ``\textbf{ATTIQA}'', \textbf{ATT}ribute-aware \textbf{IQA}, which exhibits enhanced generalization capabilities by effectively incorporating CLIP's extensive knowledge and the scalability of large unlabeled datasets.
While previous works~\cite{wang2023exploring,zhang2023blind} have observed that CLIP inherently contains robust representations relevant to IQA, the representation of CLIP also consists of a wide range of semantic contexts, hindering the precise assessment of image qualities.
To this end, we propose a pretraining scheme that distills only quality-aware knowledge from CLIP with unlabeled large dataset.
Specifically, we generate pseudo-labels for given unlabeled images utilizing CLIP's zero-shot inference with quality-aware prompts and use them for training a target encoder (\fref{fig:teaser}(c)).
Such a pretraining scheme can effectively transfer CLIP's quality-related knowledge, along with the scalability benefits of unlabeled large datasets, into the target encoder.
This results in the construction of robust representations that contain only helpful information for IQA.

To generate quality-aware pseudo-labels, we propose to incorporate prompts based on five key attributes, which have been proven to be crucial for assessing image quality~\cite{huang2022explainable, fang2020perceptual, su2021koniq++}.
Specifically, we propose a five image attribute based pretraining strategy beyond Mean Opinion Score(MOS).
Instead of using generic prompts such as ``a good/bad photo'', as used in CLIP-IQA~\cite{wang2023exploring}, we select prompts representing each key attribute through Large Language Model(LLM) and our carefully designed proxy tasks.
They are taken by CLIP as inputs to generate pseudo-labels, facilitating a network to learn from five representation spaces for each specific image attribute (\fref{fig:teaser}(c)).
Taking advantage of using a large-scale dataset combined with the novel attribute-aware CLIP guidance, our pretraining framework significantly enhances the learning of rich representations closely associated with image attributes and quality.
We demonstrate the effectiveness of our method through extensive experiments, achieving state-of-the-art performance on multiple IQA datasets, as well as on an aesthetic quality dataset.

The ability to generalize beyond the training dataset is crucial for IQA, particularly when considering its further applications.
We observe that ATTIQA exhibits superior performance when the test dataset is unseen (\fref{fig:teaser}(d)) or the training dataset is limited, which is more applicable to real-world scenarios.
Building on these strengths, we propose a couple of applications where a generalizable IQA method can be employed.
We show that our method can be used to evaluate the outputs of a generative model~\cite{rombach2022high} and as a reward function for reinforcement learning-based image enhancement \cite{shin2022drl}.

\section{Related Work}
\label{sec:related}

\textblock{Classical Image Quality Assessment.}
Since image quality is highly regarded as essential in diverse vision applications, numerous image quality assessment studies have been explored.
Traditional NR-IQA utilizes a feature-based machine learning approach to quantify image quality, leading to a primary focus on extracting meaningful features.
Therefore, these works introduced hand-crafted feature based IQA, which is derived from natural scene statistics~\cite{gao2013universal}, spatial domain~\cite{mittal2012no, mittal2012making} or frequency domain~\cite{saad2012blind}.

\textblock{Deep learning based IQA.}
With the success of deep learning, various deep learning-based NR-IQA methods have been introduced.
Early works tried to train neural networks by directly predicting mean opinion score (MOS)~\cite{kang2014convolutional} or the distribution of MOS~\cite{talebi2018nima}.
Some works attempted to incorporate meta-learning~\cite{zhu2020metaiqa} or hypernetwork~\cite {su2020blindly}.
However, the limited size of IQA datasets restricts deep learning-based approaches, making it hard to extract rich representations solely relying on the IQA dataset.
To address this problem, recent IQA methods~\cite{9271914,talebi2018nima, zhang2018blind} commonly adopted ImageNet~\cite{russakovsky2014imagenet} classification-based backbone as their initial state, which already possesses rich representations.
However, there is another problem that this representation is not fully suitable to IQA, as their pretraining task mainly focuses on semantic information but not image quality.

\textblock{Pretrain based IQA.}
Beyond applying deep learning strategies to IQA, some approaches have focused on generating quality-aware representation using large datasets without the need for ground truth.
Liu \textit{et al.}~\cite{liu2017rankiqa} employed the Siamese network to rank images according to their quality, generating images of varying quality levels by applying different scales of distortion to a single image.
Synthetic distortion-aware representation was introduced in~\cite{zhang2018blind}, which attempts to classify the type or the amount of distortions applied to images. 
Recently, with the success of SSL, some works \cite{madhusudana2022image,saha2023re,zhao2023quality} suggested a contrastive learning framework refined for IQA.
Unlike typical contrastive learning, they viewed patches from the same image, and each patch is differently distorted as different-quality samples.
By treating these samples as negative pairs in the training process, they efficiently constructed a quality-aware representation space.

\textblock{Vision Language Model based IQA.}
VLM~\cite{li2022blip, ramesh2021zero} is a foundation model that learns correspondence between image and text to understand the relationships between visual contents and language.
Specifically, CLIP~\cite{radford2021learning} is trained with 400 million image-text pairs, and thus, it shows generalization capability for various computer vision tasks.
Taking advantage of this ability, IQA methods that utilize CLIP have also been introduced.
CLIP-IQA \cite{wang2023exploring} directly assessed image quality by measuring the image's correspondence with quality-aware text prompts.
They also suggested an enhanced version named CLIP-IQA+ that optimizes text prompts to adapt to the given target dataset.
Despite this successful application, since the CLIP is trained with unrefined caption data focusing on semantic information, there is still room for improvement in refining CLIP's representation space towards an image quality-aware representation space.
In response to this property, some works have tried to adapt the representation space of CLIP with additional datasets.
Ke \textit{et al.}~\cite{ke2023vila} fine-tuned the CoCa~\cite{Yu2022CoCaCC} with the aesthetic captioning dataset to make aesthetic-aware VLM.
By injecting aesthetic-related vision-language correspondence into VLM, they showed improved performance of their representation in quality-related downstream tasks.
Zhang \textit{et al.}~\cite{zhang2023blind} suggested a multi-task learning approach to adapt the vision backbone of VLM for a unified IQA dataset.
They trained the vision backbone by optimizing cosine similarities of multi-modal embeddings with task-aware prompts.

As demonstrated by the above approaches, works refining the representation of VLM to aware image quality are currently underway.
While these methodologies showed improvements in incorporating quality-aware information into VLM's representation space, the necessity of additional datasets to fine-tune remains a limitation.
To mitigate this issue, our method does not \textit{fine-tune} the CLIP model itself, but rather \textit{extract} quality-related information from CLIP and use them to train our IQA model.
\begin{figure*}[t]
\centering
\includegraphics[width=\linewidth]{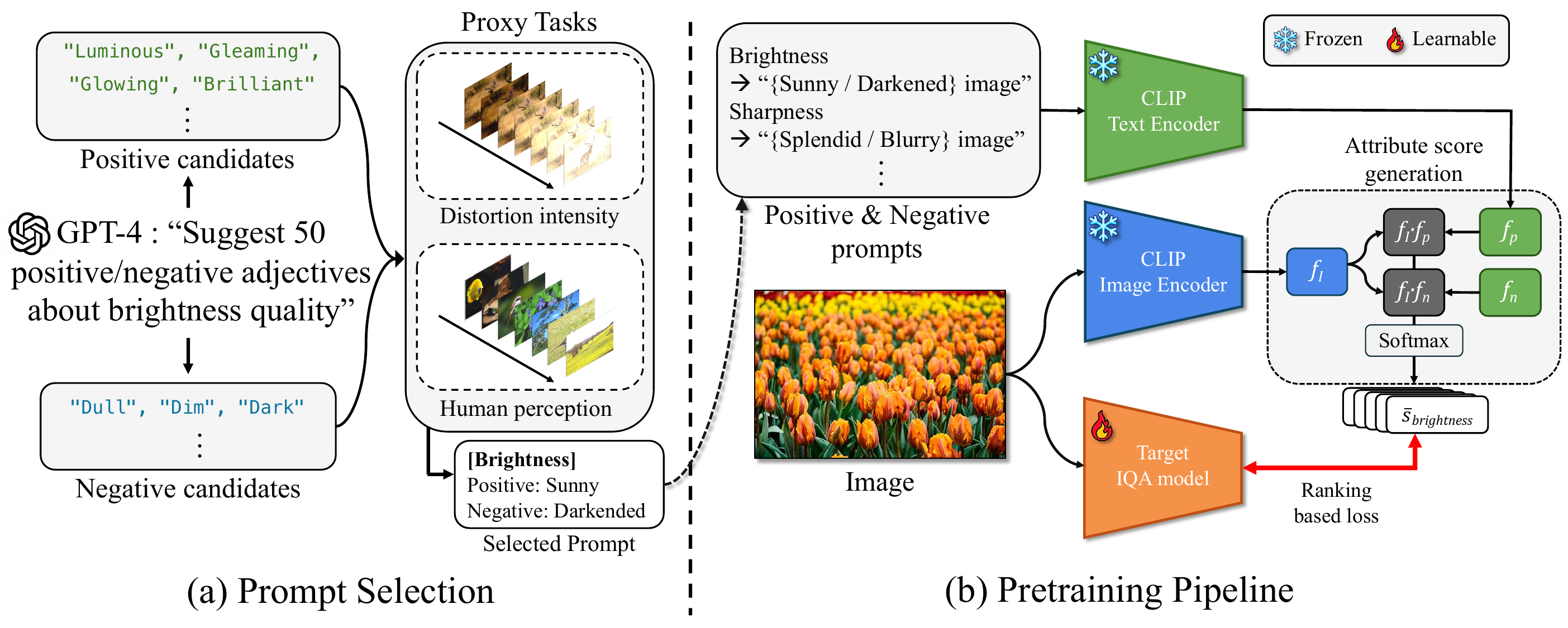}
\caption{
(a) The overall process of our prompt selection strategy for each image attribute (e.g. brightness). Given attribute, we create prompt candidates using GPT-4 and then find the optimal prompt by utilizing proxy tasks related to the attribute. (b) ATTIQA's proposed pretraining pipeline. We generate attribute scores using CLIP with an antonym strategy and then train our target IQA model using ranking-based loss with generated scores.}
\label{fig:method}
\end{figure*}

\section{Method}
\label{sec:methods}
\fref{fig:method} illustrates our framework, which consists of two primary components: prompt selection and pretraining pipeline using pseudo-labels from CLIP~\cite{radford2021learning}. 
During the prompt selection phase, we create a list of candidate prompts using a large language model (LLM). 
We then identify the most suitable prompts for generating image attribute scores by evaluating the score generation ability of candidates through proxy tasks.
In the pretrain stage, we generate image attribute scores as pseudo-labels for pretraining using CLIP and the chosen prompts.
Our IQA model is pretrained on this pseudo-labeled data and fine-tuned using a dedicated IQA dataset.

\subsection{Image Attributes}

Our method aims to utilize image attribute-based scores as supervision to reduce the ambiguity of the IQA task, which is typically represented solely by MOS.
This approach yields a more precise and well-defined representation of image quality by offering specific quality criteria beyond the generic and ambiguous mere notions of `good' and `bad'.

In the line of IQA research that aims to incorporate quality relevant information beyond the MOS, five key attributes -- \textit{sharpness, contrast, brightness, colorfulness, and noisiness}-- are widely employed and have proven beneficial for the IQA task~\cite{huang2022explainable, fang2020perceptual, su2021koniq++}.
Especially, SPAQ \cite{fang2020perceptual} substantiated through a user study that these five image attributes correlate well with perceived image quality. 
These observations indicate that these attributes are helpful factors in understanding the image quality.
Therefore, we choose these five attributes as the target objective for our IQA model.
Note that the following attribute score generation and prompt selection are conducted separately for each image attribute.

\subsection{Attribute Score Generation}
\label{sec:pseudo_label}
During attribute score generation, we generate five attribute scores for given images using CLIP's zero-shot inference.
Given image \textrm{x} and attribute-aware prompt $t$, CLIP encodes the image and prompt into shared multi-modal feature space. 
We then compute the relatedness score $s$ between \textrm{x} and $t$ using cosine similarity as follows:
\begin{equation}
    \begin{aligned}
         s(\textrm{x},t) &= {E_I(\textrm{x}) \cdot E_T(t)^T \over ||E_I(\textrm{x})|| \cdot ||E_T(t)||},
    \end{aligned} 
    \label{eq:zero-shot}
\end{equation}
where $E_I$ and $E_T$ represent CLIP's image and text encoder, respectively.

Our pseudo-label generation employs an antonym strategy~\cite{wang2023exploring}, which computes scores by integrating scores of positive and negative prompts with the softmax function.
For example, we can use a prompt pair \{``\textit{Dark image}'', ``\textit{Bright image}''\} as a negative-positive pair to calculate the brightness attribute score.
Then, our attribute score is computed by the following equation:
\begin{equation}
    \begin{aligned}
         \bar{s}_{attribute}(\textrm{x}) &= {e^{s(\textrm{x},t_{pos})} \over e^{s(\textrm{x},t_{pos})} + e^{s(\textrm{x},t_{neg})}},
    \end{aligned} 
    \label{eq:scoring}
\end{equation}
where $t_{pos}$ and $t_{neg}$ represent positive and negative prompts for the corresponding image attribute, respectively.

\subsection{Prompt Selection}
\label{sec:prompt_selection}

Previous work involving CLIP~\cite{zhou2022learning} has shown that the choice of prompts is critical in determining performance. To address this, we introduce a prompt selection strategy aimed at identifying the most effective prompts for generating image attribute scores. Drawing inspiration from techniques used in the NLP field~\cite{gao2021making}, we develop a selection based efficient approach for identifying the optimal prompts.

As depicted in the left side of Figure \ref{fig:method}(a), we begin by generating prompt candidates using Large Language Model (LLM), specifically GPT-4~\cite{openai2023gpt4}. 
To streamline the search process, we adopt a standard template for these prompts in the format \textit{``[adjective] image"}, focusing specifically on a variation of adjectives.
Prompt candidates are constructed using GPT-4, with an ask query to elicit adjectives pertinent to specific image attributes.
Since we use antonym pair for each attribute, we generate 50 positive and 50 negative adjectives, resulting in 2500 positive-negative prompt candidates for each attribute.
Subsequently, we identify the most suitable prompt pair from 2500 candidates by assessing their capability to generate accurate attribute scores.
To find the optimal prompt pair, we present two proxy tasks.
The optimal prompt is determined as the prompt that produces an attribute score that best aligns with the goals of both tasks.
To measure an image attribute appropriately, the proxy tasks are designed under two hypotheses: 1) For a fixed image, when a distortion corresponding to the image attribute is applied to it, the attribute score predicted by the prompt should increase or decrease accordingly. 2) For different images, the attribute scores generated by the prompt pairs should match well with the degree of human perception of each attribute.
\begin{wraptable}{r}{0.45\columnwidth}
\label{t3}
\centering
    \caption{Results of the prompt selection. These prompts are chosen by our prompt selection strategy.}
    \resizebox{0.45\columnwidth}{!}
    {
            \begin{tabular}{c||cc}
        \toprule
        Attribute & Positive prompt & Negative prompt \\
        \midrule
        Sharpness & \textit{"Splendid image"} & \textit{"Blurry image"}\\
        Contrast  & \textit{"Distinct image"} & \textit{"Vague image"}\\
        Brightness & \textit{"Sunny image"} & \textit{"Darkened image"}\\
        Colorfulness & \textit{"Vibrant image"} & \textit{"Colorless image"} \\
        Noisiness & \textit{"Peerless image"} & \textit{"Grainy image"} \\
        \bottomrule
        \end{tabular}
    }
\label{table:method_prompt}
\end{wraptable}

In the first proxy task, we apply varying levels of distortion to a fixed image and identify the optimal pair of prompts that yield an attribute score aligning most accurately with the applied level of distortion.
For the second task, we employ the SPAQ dataset, which comprises diverse scenes and offers human-annotated attribute scores for the same five attributes we adopt. We aim to identify the prompt pair that generates the attribute scores whose order closely aligns with the order of the provided scores in the dataset.
We calculate the sum of SROCC scores for both tasks and select the highest performing prompt pair, and the result is shown in \Tref{table:method_prompt}.
These selected pairs are then utilized to generate each attribute score in the following pretraining pipeline.

\subsection{Attribute Aware Pretraining Pipeline}
\label{sec:training_pipeline}
After selecting prompts, we train the target IQA model to construct attribute-aware space with ranking-based loss using a pseudo-label derived from CLIP, as illustrated in \fref{fig:method}(b).

Our method aims to create five unique representation spaces for each specific image attribute.
Accordingly, our IQA model comprises a shared encoder backbone and five attribute heads for each image attribute.
Each attribute head consists of two-layer multi-layer perceptrons (MLPs) that output an attribute score.
Then, our training objective for the pretraining pipeline can be formulated by minimizing the discrepancy between five attribute score predictions from the IQA model and the corresponding image attribute scores generated from CLIP.

Our pretraining pipeline can be directly implemented using regression-based loss such as MSE or L1 loss.
However, directly using the attribute score with regression-based loss hinders handling uncertainties.
Since scoring image attributes as scalars in a zero-shot manner is inherently challenging, training by predicting these scores may impede the construction of robust representations.
To address this problem, we use a relative ranking-based loss instead of a numerical norm-based loss to minimize the dependence on CLIP's numerical results, which are subject to uncertainty.
To implement this loss in our framework, we utilize margin-ranking loss that optimizes the relative ranking of the two samples given.
We first define the indicator function $F$, which specifies the superiority of image attribute based on its score $\bar{s}$ for given images:

\begin{equation}
    \begin{aligned}
        F_a(\textrm{x}_1,x_2) = \begin{cases} 
        0,& \bar{s}_{a}(\textrm{x}_1) > \bar{s}_{a}(\textrm{x}_2) \\
        1,& \bar{s}_{a}(\textrm{x}_1) \leq \bar{s}_{a}(\textrm{x}_2),
        \end{cases}
    \end{aligned} 
    \label{eq:matrix}
\end{equation}

where $a$ denotes an element of image attributes set $A=$\{\textit{sharpness, contrast, brightness, colorfulness, noisiness}\}, and $\textrm{x}_1$ and $\textrm{x}_2$ denote the sample images.

Then, we train our target IQA model based on margin-ranking loss with the indicator function $F_{a}$.
We compute our loss by summation of margin-ranking loss independently for each attribute $a$ using attribute score prediction from respective attribute head $E_{a}$:
\begin{equation}
    \resizebox{.6\hsize}{!}{
        $L =  \sum_{a} \max({0, m - (E_{a}(\textrm{x}_1) - E_{a}(\textrm{x}_2)))}\cdot F_a(\textrm{x}_1,\textrm{x}_2),$
        }
    \label{eq:loss}
\end{equation}
where $m$ denotes the margin hyper-parameter.
\subsection{Fine-tuning}
\label{sec:pseudo_label}
To predict MOS with our IQA model, we have to fine-tune it on the target IQA dataset.
However, the architecture of our IQA model during the pretraining stage is not designed to output a single score but instead predicts five attribute scores.
Therefore, we adapt the architecture of our model to fit the IQA task to generate a single MOS as output.
At a fine-tuning stage, we extract features from each attribute head, excluding the final layer, and these features are then concatenated and fed into regression MLPs that predict the MOS.
\section{Experiments}
\label{sec:experiments}
\subsection{Datasets}
We conduct experiments with ATTIQA on the 4 ``in-the-wild" NR-IQA Datasets, CLIVE \cite{ghadiyaram2015massive}, KonIQ-10k \cite{hosu2020koniq}, SPAQ \cite{fang2020perceptual}, and FLIVE \cite{ying2020patches} and 1 image aesthetic dataset, AVA \cite{murray2012ava}. 
While AVA\cite{murray2012ava} focuses on image aesthetics, we utilize this dataset since its user study setting is the same as ``IQA in the wild''.


For FLIVE \cite{ying2020patches} and AVA \cite{murray2012ava}, we follow the official dataset split.
For the rest, we randomly partition the dataset and allocate 80\% to the train set and 20\% to the test set.
Following previous works~\cite{zhao2023quality}, we conduct the same experiment ten times with different random splits and report the median value as a result to compensate for the bias arising from random splits. 

\subsection{Experimental Setup}
For a fair comparison, we utilize ResNet-50~\cite{he2016resnet} as our backbone, widely used in NR-IQA.
For CLIP, we adopt the ViT-B/16 model\cite{dosovitskiy2021an}.
We experimentally set the value of the margin parameter $m$ at the loss function to 0.1.

At the pretraining stage, we use the ImageNet~\cite{russakovsky2014imagenet} widely used for pretext tasks.
At the fine-tuning stage, we followed the setting from~\cite{zhao2023quality}. 
We resized the image's shorter edge to 340 and randomly cropped the image at a resolution of 320$\times$320.
We fine-tuned our network to 100 epochs on each target dataset.
At the evaluation stage, we take five crops at a resolution of 320$\times$320 from each corner and center as test samples, and the average of the results is used for the predicted MOS.
For performance evaluation, we calculated Pearson’s Linear Correlation Coefficient (PLCC) and Spearman’s Rank-Order Correlation Coefficient (SROCC), widely adopted evaluation metrics in IQA research.
\begin{table*}[t]
\label{t3}
\begin{center}
\caption{Fine-tuning performance comparison of ATTIQA and existing NR-IQA methods for 4 IQA ``in-the-wild'' dataset and 1 IAA dataset. ``$\dag$'' denotes that this measurement is achieved from \cite{zhao2023quality}. ``-'' denotes that measurement is not possible due to the absence of an official code and result. Other measurements are based on the official reports or reproduced by the official code. We highlight the best performance in bold and underline the second-best performance for each dataset.}
\resizebox{\linewidth}{!}
{
    \begin{tabular}{c||cc|cc|cc|cc|cc}
    \toprule
    \multirow{2}{*}{Methods} & \multicolumn{2}{c}{CLIVE} & \multicolumn{2}{c}{KonIQ} & \multicolumn{2}{c}{SPAQ} & \multicolumn{2}{c}{FLIVE} & \multicolumn{2}{c}{AVA} \\
    & SROCC & PLCC & SROCC & PLCC & SROCC & PLCC & SROCC & PLCC & SROCC & PLCC \\
    \midrule
    DBCNN$^\dag$ \cite{zhang2018blind} & 0.844 & 0.862 & 0.878 & 0.887 & 0.906 & 0.907 & 0.542 & 0.626 & 0.554 & 0.583 \\
    HyperIQA$^\dag$~\cite{su2020blindly} & 0.855 & 0.871 & 0.908 & 0.921 & 0.916 & 0.919 & 0.535 & 0.623 & 0.668 & 0.668\\
    CONTRIQUE~\cite{madhusudana2022image}& 0.824 & 0.848 & 0.900 & 0.915 & 0.910 & 0.915 & 0.598 & 0.674 & 0.674 & 0.678\\
    MUSIQ~\cite{ke2021musiq} & - & - & 0.916 & 0.928 & 0.917 & 0.921 & \textbf{0.646} & 0.739 & 0.726 & 0.738\\
    TReS~\cite{golestaneh2022no} & 0.846 & 0.877 & 0.915 & 0.928 & 0.915 & 0.919 & 0.554 & 0.625 & 0.658 & 0.663 \\
    Re-IQA~\cite{saha2023re} & 0.823 & 0.865 & 0.924 & 0.935 & 0.915 & 0.919 & 0.574 & 0.674 & 0.714 & 0.716\\
    QPT~\cite{zhao2023quality}& \underline{0.895} & 0.914 & 0.927 & 0.941 & \underline{0.925} & \underline{0.928} & 0.610 & 0.677 & -& -\\
    CLIP~\cite{radford2021learning}& 0.847 & 0.881 & 0.918 & 0.932 & 0.918 & 0.922 & 0.563 & 0.628 & 0.746 & 0.745 \\
    CLIP-IQA+~\cite{madhusudana2022image} & 0.805 & 0.832 & 0.895 & 0.909 & 0.864 & 0.866 & 0.575 & 0.593 & 0.692 & 0.732 \\
    LIQE~\cite{zhang2023blind}&0.865 & 0.866 & 0.898 & 0.913 & - & - & - & - & -& - \\
    \midrule
    ATTIQA (Distortion intensity) & 0.891 & 0.910 & \underline{0.929} & \underline{0.943} & 0.922 & 0.926 & 0.625 & 0.729 & 0.754 & 0.750\\
    ATTIQA (Human perception) & 0.890 & \underline{0.915} & \textbf{0.942} & \textbf{0.952} & \underline{0.925} & \textbf{0.930} & \underline{0.635} & \underline{0.740} & \underline{0.756} & \underline{0.759} \\
    ATTIQA (Joint strategy) & \textbf{0.898} & \textbf{0.916} & \textbf{0.942} & \textbf{0.952} & \textbf{0.926} & \textbf{0.930} & 0.632 & \textbf{0.742} & \textbf{0.761} & \textbf{0.761} \\
    \bottomrule
    \end{tabular}
}
\label{table:main}
\end{center}
\end{table*}

\subsection{Main Result}
In this section, we report the quantitative performance of ATTIQA and compare it with existing NR-IQA models.
Utilizing our CLIP-guided attribute-aware pre-trained model, we conduct fine-tuning on five IQA datasets to predict the MOS.
As shown in the \Tref{table:main}, ATTIQA shows notable performance improvements compared to CLIP-based methods on four IQA ``in the wild'' datasets and one image aesthetic dataset AVA.
Our method demonstrates state-of-the-art performance in most evaluation settings, with the second-best SROCC performance on the FLIVE dataset.
It is important to note that MUSIQ is an exceptional work that utilized the complete FLIVE dataset comprising patches and full images, unlike the other methods that do not use patch data.
Notably, ATTIQA shows a significant performance gap on the KonIQ-10k and AVA datasets.

In the last three rows of \Tref{table:main}, we report the performance of three different versions of ATTIQA. 
We carry out experiments with various types of prompts, including cases where we apply the two proxy tasks described in Sec \ref{sec:prompt_selection}—Distortion Intensity and Human Perception—separately, as well as a scenario where we combine both tasks in our prompt selection strategy (Joint Strategy).
We observe that prompts based on \textit{human perception} work effectively, and the \textit{joint strategy} that involves both proxy tasks shows the most superior performance.
It indicates that a prompt selection strategy that considers using both low-level information \textit{distortion} and high-level \textit{human perception} enhances the robustness of our model across various datasets.
\subsection{Generalization Capability}

\begin{table}[t]
\caption{Cross dataset validation performance comparison of ATTIQA and existing NR-IQA methods.}
\label{t3} 
\resizebox{\columnwidth}{!}
{
    \centering
    \begin{tabular}{c||ccc|ccc|ccc|ccc}
    \toprule
     Train DB & \multicolumn{3}{c}{CLIVE} & \multicolumn{3}{c}{KonIQ} & \multicolumn{3}{c}{SPAQ} & \multicolumn{3}{c}{FLIVE}\\
    \midrule
     Test DB & KonIQ & SPAQ & FLIVE & CLIVE & SPAQ & FLIVE & CLIVE & KonIQ & FLIVE & CLIVE & KonIQ & SPAQ\\
    \midrule
    CONTRIQUE & 0.676 & 0.842 & 0.346 & 0.731 & 0.789 & 0.410 & 0.549 & 0.646 & 0.338 & 0.706 & 0.709 & 0.734 \\
    Re-IQA & 0.769 & 0.852 & 0.424 & 0.791 & 0.862 & 0.461 & 0.732 & 0.707 & 0.497 & 0.720 & 0.676 & 0.793\\
    CLIP & 0.725 & 0.850 & 0.405 & 0.799 & 0.837 & 0.507 & 0.773 & \underline{0.753} & 0.496 & 0.727 & 0.717 & 0.834 \\
    CLIP-IQA+ & 0.697 & 0.836 & 0.437 & 0.803 & 0.832 & 0.516 & 0.784 & 0.722 & 0.470 & 0.620 & 0.631 & 0.661 \\
    LIQE & \underline{0.819} & \underline{0.877} & \underline{0.497} & \underline{0.824} & \underline{0.868} & \textbf{0.551} & - & - & - & - & -& -\\
    \midrule
    ATTIQA & \textbf{0.829} & \textbf{0.887} & \textbf{0.511} & \textbf{0.856} & \textbf{0.879} & \underline{0.540} & \textbf{0.824} & \textbf{0.819} & \textbf{0.548} & \textbf{0.756} & \textbf{0.762} & \textbf{0.867}\\
    \bottomrule
    \end{tabular}  
}
\label{table:cv} 
\end{table}
\noindent\textbf{Cross-dataset Validation.}
\label{sec:cv}
To verify ATTIQA's generalization ability, we conduct experiments about cross-dataset validation.
This experiment evaluates the IQA model's ability to learn generalizable features by training it on the specified dataset and testing it on the unseen dataset.
To consider various scenarios, we conduct extensive experiments across four datasets: CLIVE, KonIQ, SPAQ, and FLIVE.
Every experimental setup is the same as the main experiment, and due to the various ranges of the MOS for each dataset, we use only SROCC as an evaluation criterion.
As shown in \Tref{table:cv}, ATTIQA exhibits superior generalization capability to baselines, achieving the best performance in most scenarios.
Interestingly, LIQE achieves comparable results to ATTIQA, demonstrating that strategies adapting CLIP possess strong generalization capabilities.
However, we highlight that ATTIQA outperforms LIQE in most scenarios and that LIQE cannot be extended to datasets where additional annotations are not provided.

\noindent\textbf{Data-Efficient Setup.}
Moreover, we conducted experiments in a data-efficient setup to demonstrate that ATTIQA can generalize in environments where only a small amount of data is available.
Instead of the conventional 8:2 Train-Test split, we performed training using only 10\% or 20\% of the data. 
Since we utilize a small amount of data, we also use only SROCC as an evaluation criterion.
As shown in \Tref{table:ld}, ATTIQA outperforms other pretrain-based methods in environments with limited datasets.
This performance gap validates that our pretraining strategy is more robust than other baselines.

\begin{table}[htb]
    \centering
    \begin{minipage}{0.54\columnwidth}
        \centering
       \caption{Comparison of ATTIQA and NR-IQA methods which focuses on representation learning under data efficient setup.}

       \resizebox{\columnwidth}{!}{
            \begin{tabular}{c||cc|cc|cc}
            \toprule
            \multirow{2}{*}{Methods} & \multicolumn{2}{c}{CLIVE} & \multicolumn{2}{c}{KonIQ} & \multicolumn{2}{c}{SPAQ}\\
            & 10\% & 20\% & 10\% & 20\% & 10\% & 20\% \\
            \midrule
            CONTRIQUE & \underline{0.687} & \underline{0.740} & 0.832 & 0.835 & 0.883 & 0.885 \\
            Re-IQA  & 0.632 & 0.683 & \underline{0.853} & \underline{0.888} & \underline{0.893} & \underline{0.902} \\
            CLIP & 0.650 & 0.728 & 0.846 & 0.863 & 0.882 & 0.884 \\
            \midrule
            ATTIQA & \textbf{0.820} & \textbf{0.838} & \textbf{0.903} & \textbf{0.919}  & \textbf{0.909} & \textbf{0.917} \\
            \bottomrule
            \end{tabular}
        }
        \label{table:ld}
    \end{minipage}
    \begin{minipage}{0.435\columnwidth}
        \label{t3} 
        \centering
        \caption{
        Cosine similarity between features from pretrained and fine-tuned encoder.}
        \resizebox{\columnwidth}{!}
        {
            \centering
            \begin{tabular}{c||c|c|c}
            \toprule
             Fine-tune DB & CLIVE & KonIQ & SPAQ\\
            \midrule
            \makecell{CONTRIQUE} & 0.536 & 0.566 & 0.613\\
            \makecell{Re-IQA} & 0.158 & 0.181  & 0.243\\
            \makecell{CLIP} & 0.195 & 0.203  & 0.309\\
            \midrule
            \makecell{ATTIQA} & \textbf{0.890} & \textbf{0.811} & \textbf{0.945}\\
            \bottomrule
            \end{tabular}  
        }
        \label{table:feature} 
    \end{minipage}
\label{table:minipage}
\end{table}
\noindent\textbf{Feature Analysis.}
In this section, we analyze why ATTIQA exhibits superior generalization capability compared to other pretrain-based methods.
We hypothesize that pretext tasks providing more generalizable representations would offer robust features that do not overfit specific datasets.
To examine these properties, we extract features from each backbone before and after fine-tuning the IQA dataset and compare them by measuring cosine similarity.
As shown in \Tref{table:feature}, ATTIQA's features are slightly adjusted, whereas other methods' features are modified significantly.
This result suggests that ATTIQA's pretrained representation inherently possesses superior robustness and provides a more effective initialization point for IQA than other methods, leading to enhanced performance and generalization capability.

For real-world applications, a model's generalization ability is far more critical than its performance on specific benchmark datasets.
Our method's superior generalization capability ensures robust baseline performance on unseen images, highlighting its practicability when considering the purpose of the IQA tasks.
Building on this strength, we will demonstrate the application of ATTIQA in real-world scenarios in Sec \ref{sec:application}.

\subsection{Ablation Studies}

\textblock{Linear probing.}
We conduct linear probing experiments to demonstrate the robustness of our attribute-aware pretrained feature space, training only a single regression MLPs on the target dataset with a frozen ATTIQA backbone.
In this experiment, we compared ATTIQA to previous works that suggest pretext tasks for IQA.
For Re-IQA, we report the three types of results based on the encoder configuration: using only the quality or content encoder, and both encoders.
As shown in \Tref{table:lp}, ATTIQA shows a significant performance gap compared to other methods in CLIVE.
In other datasets, ATTIQA also demonstrates comparable results to Re-IQA, which uses both features of a separate quality and content encoder, while our method only utilizes a single shared encoder for five attributes.
We note that CLIP shows the worst performance, validating our motivation that CLIP's original representations are unsuitable for precise IQA.

\begin{table}[htb]
    \begin{minipage}{0.495\columnwidth}
        \centering
       \caption{Linear probing performance comparison of ATTIQA and NR-IQA methods which focuses on representation learning.}
       \resizebox{\columnwidth}{!}{
            \begin{tabular}{c||cc|cc|cc}
            \toprule
            \multirow{2}{*}{Methods} & \multicolumn{2}{c}{CLIVE} & \multicolumn{2}{c}{KonIQ} & \multicolumn{2}{c}{SPAQ}\\
            & SROCC & PLCC& SROCC & PLCC& SROCC & PLCC \\
            \midrule
            CONTRIQUE & \underline{0.845} & \underline{0.857} & 0.894 & 0.906 & \underline{0.916} & 0.919 \\
            Re-IQA (quality) & 0.806 & 0.824 & 0.861 & 0.885 & 0.900 & 0.910 \\
            Re-IQA (content) & 0.808 & 0.844 & 0.896 & 0.912 & 0.902 & 0.908 \\
            Re-IQA (both) & 0.840 & 0.854 & \textbf{0.914} & \textbf{0.923} & \textbf{0.918} & \textbf{0.925} \\
            CLIP & 0.803 & 0.829 & 0.883 & 0.895 & 0.895 & 0.896 \\
            \midrule
            ATTIQA & \textbf{0.870} & \textbf{0.891} & \underline{0.903} & \underline{0.918}  & \textbf{0.918} & \underline{0.922} \\
            \bottomrule
            \end{tabular}
        }
        \label{table:lp}
    \end{minipage}
    \hspace{1mm}
    \begin{minipage}{0.49\columnwidth}
        \centering
        \caption{Ablation study results about our prompt based strategy and loss function.}
        \resizebox{\columnwidth}{!}{
            \begin{tabular}{c||cc|cc|cc}
        \toprule
        \multirow{2}{*}{Prompt type}& \multicolumn{2}{c}{CLIVE} & \multicolumn{2}{c}{KonIQ} & \multicolumn{2}{c}{SPAQ}\\
        & SROCC & PLCC& SROCC & PLCC& SROCC & PLCC \\
        \midrule
        ATTIQA & \textbf{0.898} & \textbf{0.916} & \textbf{0.942} & \textbf{0.952} & \textbf{0.926} & \textbf{0.930} \\
        \midrule
        single-prompt & 0.880 & 0.909 & 0.928 & 0.939 & 0.916 & 0.920 \\
        \midrule
        Worst prompt & 0.869 & 0.889 & 0.930 & 0.943 & 0.920 & 0.925 \\
        Median prompt & 0.872 & 0.893 & 0.931 & 0.944 & 0.921 & 0.925 \\
        \midrule
        with $L_2$ &  0.875 & 0.904 & 0.933 & 0.945 & 0.923 & 0.928 \\  
        \bottomrule
        \end{tabular}
        }
        \label{table:ablation}
    \end{minipage}
    \vspace{-5mm}
\label{table:minipage}
\end{table}
\textblock{Attribute based Approach.}
\label{sec:attribute}
To demonstrate the effectiveness of our image attribute based approach, we carry out an experiment by replacing the target objective from five image attributes with a single overall image quality.
In this experiment, we train the IQA model with a single pseudo-label using a prompt pair that describes image quality: \textit{\{``Good image", ``Bad image"\}}.
Comparing the first and the second tab of \Tref{table:ablation}, we can observe that our representation space decomposing image quality into five attributes outperforms the single-prompt based representation space, justifying our approach for model design.

\textblock{Prompt Selection Strategy.}
To justify our prompt selection strategy, we experiment with other prompts selected by different strategies.
For comparison, we adopt prompts that achieve the median and lowest scores in the proxy task.
As shown in the third tab of \Tref{table:ablation}, the results of our strategy align with the performance at the evaluation. 
This correlation validates the efficacy of our prompt selection strategy.

\textblock{Ranking-based loss.}
To verify the efficacy of our relative ranking-based loss approach, we conduct an additional ablation study by replacing the margin-ranking loss with L2 loss at the pretraining stage.
As depicted in the last row of \Tref{table:ablation}, the use of L2 loss exhibits a performance degradation compared to adopting margin-ranking loss.
Interestingly, we can observe a notable performance decline in the CLIVE dataset, which has the smallest dataset size within this ablation study.
This result supports the use of relative loss instead of numerical loss, enhancing our pretraining pipeline's robustness.

\section{Applications}
\label{sec:application}

\begin{table}[htb]
\centering
    \begin{minipage}{0.53\columnwidth}
        \centering
        \captionof{table}{Comparisons of accuracy between human preferences and IQA model's result.} 
        \resizebox{\columnwidth}{!}{
            \begin{tabular}{c||c|c|c|c}
            \toprule
            Method & CONTRIQUE & Re-IQA & CLIP-IQA+ & ATTIQA \\
            \midrule
            Accuracy (\%) & 61.5 & 55.0 & 57.5 & \textbf{71.0} \\
            \bottomrule
            \end{tabular}
        }
        \label{table:diffusion}
        \captionof{table}{Performance comparisons among IQA model in an AI-Generated Contents Dataset(AGIQA-3k)} 
        \resizebox{\columnwidth}{!}{
            \begin{tabular}{c||c|c|c|c}
            \toprule
            Method &  CONTRIQUE & Re-IQA & CLIP-IQA+ & ATTIQA \\
            \midrule
            SROCC & 0.643 & 0.807 & 0.835 & \textbf{0.854} \\
            PLCC & 0.795 & 0.876 & 0.885 & \textbf{0.911} \\
            \bottomrule
            \end{tabular}
        }
        \label{table:agiqa}
    \end{minipage}
    \begin{minipage}{0.41\columnwidth}
        \centering
        \includegraphics[clip, trim={1 0 0 0}, width=0.95 \textwidth]{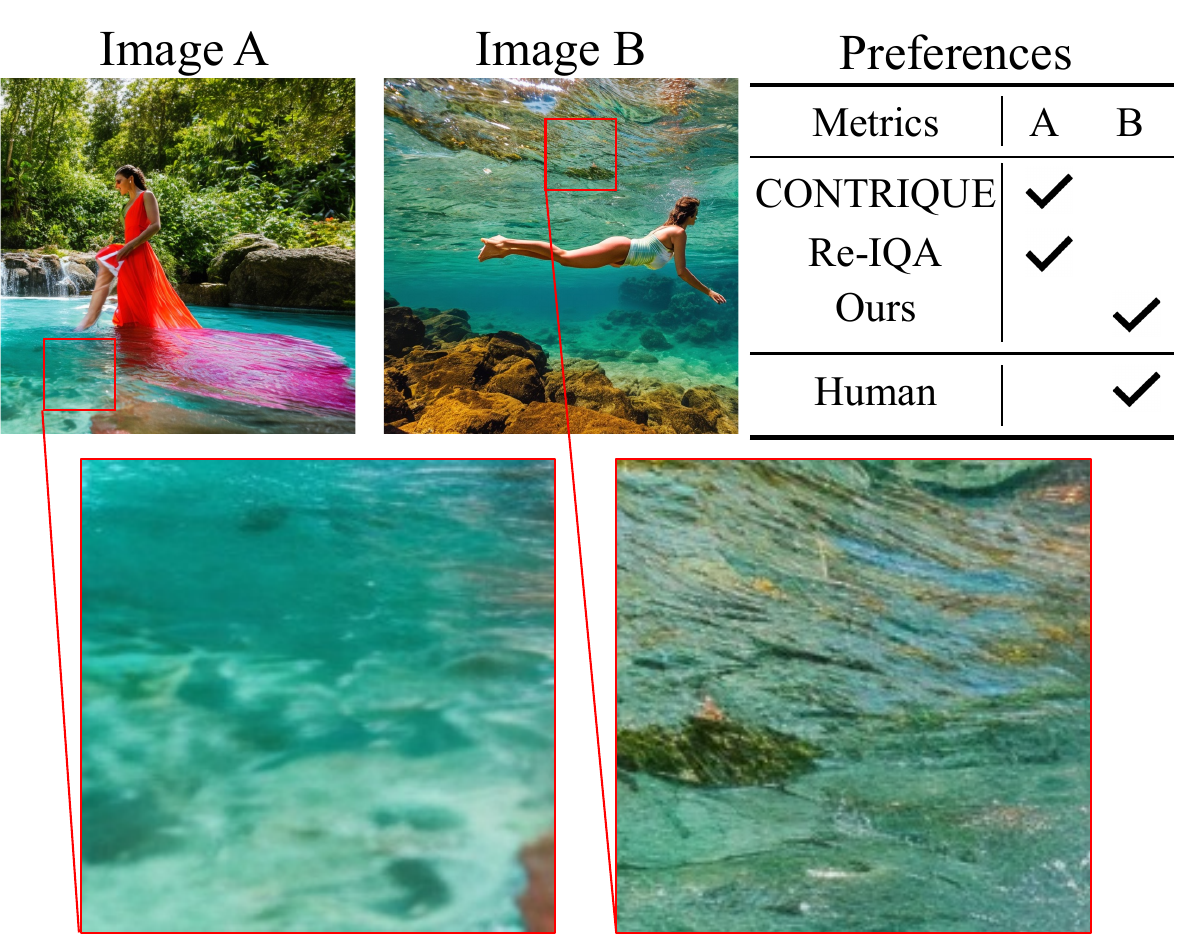}
                \captionof{figure}{Example of generated images. The images are generated by the same prompt. ATTIQA hits human preference while others do not.}
        \label{figure:diffusion}
    \end{minipage}
\end{table}

To better demonstrate our ATTIQA's generalization capability, we introduce two types of applications in this section: (1) metrics for the generative model and (2) image enhancement guided by our IQA score.
For each application, we employ models trained on the KonIQ dataset, which shows the best generalization capability at Sec \ref{sec:cv}.

\subsection{Metrics for Generative Model}
Recently, as the diffusion models~\cite{ho2020denoising,rombach2022high} have shown success in the text-to-image generation~\cite{Nichol2021GLIDETP, ramesh2021zero} task. 
One of their primary focus is generating high quality images from a given text prompt.
In this regard, we attempt to employ ATTIQA as a metric for generative models.

To validate ATTIQA's effectiveness as a metric, we create a benchmark dataset that involves the pairwise comparison of two images generated from the same text prompt.
Here, we generate 200 pairs of images using the Stable Diffusion~\cite{rombach2022high} and collect human preference by conducting a user study.
When collecting the user preferences, we only present the generated images without the prompt to make participants focus on visual quality.
The user study was carried out with 60 participants through Amazon Mechanical Turk (AMT).
We then investigate the correlation between IQA models and the human participants.

As shown in \Tref{table:diffusion} and \Fref{figure:diffusion}, our method mostly aligns with human preference compared to other IQA methods.
Our ATTIQA can capture this detailed visual quality difference while others do not.
Please refer to the supplementary for the user study details and more visual results.
We will make the benchmark used in this application publicly available for further IQA research.

Moreover, we carry out an additional experiment using an AGIQA-3k dataset \cite{agiqa}, which consists of images generated by various generative models.
As shown in \Tref{table:agiqa}, ATTIQA outperforms other methods, exhibiting a significant performance gap.
These results highlight the improved generalization capability of our method when extended to AI-generated content. 
They indicate the potential for expanding the use of ATTIQA as a metric to evaluate generative models.

\subsection{Image Enhancement}
The image signal processing pipeline (ISP) converts an input raw image into a color image.
It is essential to carefully tune the parameters of the ISP to obtain visually pleasing images.
In this section, we apply ATTIQA's MOS prediction as a reward for reinforcement learning to find optimal parameters for the ISP \cite{shin2022drl}.
After the training, we convert raw images into color images in the MIT-Adobe-5k dataset \cite{fivek}, which consists of 5,000 raw images and color images retouched by five experts (A/B/C/D/E).
Then, we conduct a user study comparing our result against the retouched one by expert C, which is typically used as the ground truth in most previous image enhancement research.
The study was executed with 60 participants through AMT, involving a comparison of 200 image pairs.
For details on the implementation, please refer to the supplementary materials.

\begin{figure}[t]
\centering
\includegraphics[clip, trim={0 0 20 0}, width=1.0 \textwidth]{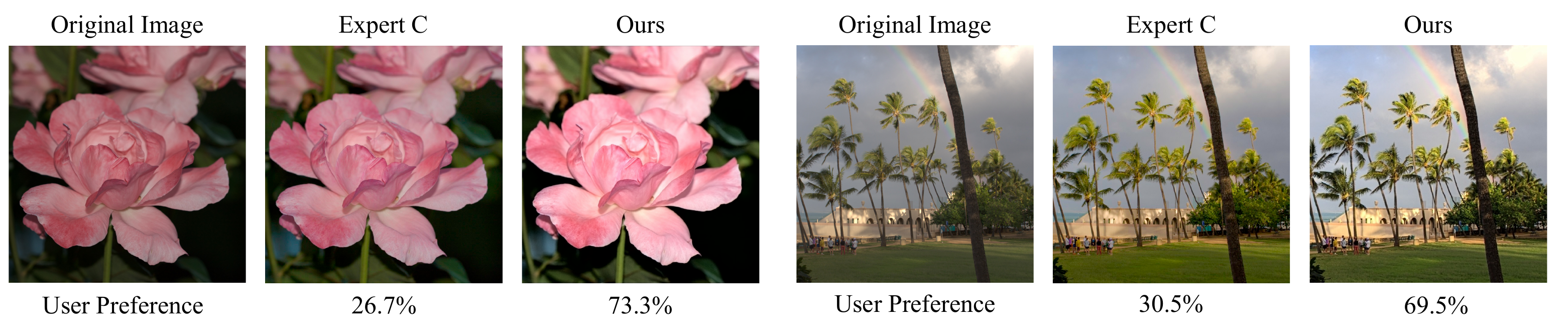}
\caption{Qualitative comparisons between our enhancement method and retouching of Expert C. Our results give more liveliness and vibrancy, aligned more closely with human preference.
}
\vspace{-5mm}
\label{fig:enhancement}
\end{figure}

As shown in \fref{fig:enhancement}, our pipeline retouches images to make them more colorful and vivid compared to both retouching by expert C and the default settings.
Furthermore, according to our user study, ATTIQA receives higher preferences from subjects, demonstrating a 58\% win rate compared to Expert C.
We also report additional qualitative comparisons to supplementary material.

\section{Discussion and Conclusion}
\label{sec:conclusion}
We propose ATTIQA, a pretraining framework for IQA that develops an attribute-aware representation space with CLIP guidance.
Since our IQA model effectively incorporates CLIP's vast knowledge and scalability of large datasets, it shows state-of-the-art performance on IQA datasets and superior generalization capability on cross-dataset validation.
Leveraging these advantages, we successfully demonstrate a couple of real-world applications where IQA can be utilized.

\textblock{Limitation and Future Work.}
While our approach focuses on five attributes commonly employed in the IQA domain, we expect that other properties relevant to image quality exist (e.g., Composition and Focus). 
Consequently, future work will involve exploring extended representation spaces for IQA. 
Given that our method proposes a pretraining framework not limited to the specified attributes, our work holds the potential for expansion to encompass additional properties.

\section*{Acknowledgements} 
This work was supported by Institute of Information \& communications
Technology Planning \& Evaluation (IITP) grant funded by the Korea government(MSIT) (Artificial Intelligence Graduate School Program, Yonsei University, under Grant 2020-0-01361).

%
%
\bibliographystyle{splncs04}
\bibliography{main}
\end{document}


\title{(Supplementary Material) \\ ATTIQA: Generalizable Image Quality Feature Extractor using Attribute-aware Pretraining} 

\titlerunning{ATTIQA}

\author{Daekyu Kwon\inst{1}\orcidlink{0009-0006-2154-3348} \and
Dongyoung Kim\inst{1}\orcidlink{0009-0000-6414-2380} \and
Sehwan Ki\inst{2}\orcidlink{0000-0002-3809-7886} \and
Younghyun Jo\inst{2}\orcidlink{0000-0002-8530-9802} \and \\
Hyong-Euk Lee\inst{2} \and
Seon Joo Kim\inst{1}\orcidlink{0000-0001-8512-216X}}

\authorrunning{Kwon et al.}

\institute{Yonsei University \and
Samsung Advanced Institue of Technology}

\maketitle

\setcounter{table}{7}
\setcounter{figure}{4}
\appendix
For supplementary materials, we provide details on our implementation and additional experiment results. 
Additionally, we attach a supplementary video that explains the overall framework of ATTIQA and the image enhancement pipeline.

\section{Details on Prompt Selection}
\label{sec:prompt_selection}
This section explains the prompts and techniques used in the prompt selection strategy. Our prompt selection strategy generates prompt candidates from GPT-4~\cite{openai2023gpt4} through the following query that receives a set of adjectives: \textit{``Suggest 50 positive/negative adjectives about \{attribute\} related to image quality"}. We create prompt candidates by changing \textit{\{attribute\}} in the query to aligned attribute. The generated prompt candidates are reported in \Tref{table:prompt_candidates}.

We also provide details on the proxy tasks stage. For the distortion intensity based proxy task, which measures attribute score by predicting the intensity of the corresponding distortion, we pair image attributes and corresponding distortions as follows:

\begin{itemize}
    \item Sharpness : Gaussian Blur, ZoomBlur, LensBlur
    \item Contrast : Contrast adjustment multiplying factors to RGB value
    \item Brightness : V adjustment in HSV space
    \item Colorfulness : Saturation adjustment in HSV space
    \item Noisiness : Gaussian Noise, ISO Noise
\end{itemize}

In attributes where multiple distortions are described, we execute proxy tasks for each distortion and compute the final result by averaging each output. In \Tref{table:proxy_prompt}, we note selected prompts that only utilize a single proxy task whose results are reported in Table \textcolor{red}{2}.

\begin{table}[t]

\label{t3}
\begin{center}
\caption{Results of the prompt selection with single proxy task. We only denote adjective for this table.}
\resizebox{0.8\columnwidth}{!}
{
    \begin{tabular}{c||cc|cc}
    \toprule
    Proxy task & \multicolumn{2}{c}{Distortion intensity} & \multicolumn{2}{c}{Human perception}\\
    \midrule
    Attribute & Positive & Negative & Positive & Negative \\
    \midrule
    Sharpness & \textit{"Unambiguous"} & \textit{"Vague"}& \textit{"High-definition"} & \textit{"Out-of-focus"}\\
    Contrast  & \textit{"Enhanced"} & \textit{"Bleak"} & \textit{"Splendid"} & \textit{"Blurred"}\\
    Brightness & \textit{"Clear"}  & \textit{"Starless"}  & \textit{"Dark"}  & \textit{"High-key"}  \\
    Colorfulness & \textit{"Multicolored"}  & \textit{"Grayish"} & \textit{"Lively"}  & \textit{"Blurred"} \\
    Noisiness & \textit{"Clutter-free"}  & \textit{"Spattered"} & \textit{"Peerless"}  & \textit{"Blurry"} \\
    \bottomrule
    \end{tabular}
}
\label{table:proxy_prompt}
\end{center}
\vspace{-8mm}
\end{table}

\section{Details on Experimental Configuration}
During pretraining, we randomly cropped the image at a resolution of 224$\times$224.
We train our network for 100 epochs, using AdamW optimizer with batch size 256 and learning rate 1e-4.

\section{Additional Experiments}
\subsection{Impact of dataset scale}
To verify the scalability of our pretraining scheme, we conduct experiments changing the ratio of the used dataset. In this experiment, we pretrain our model only utilizing 20\% and 50\% of the ImageNet dataset. As shown in \Tref{table:dataset_scale}, the performance of ATTIQA increases with the amount of available datasets. This result indicates that the performance of ATTIQA can be improved when we can train it on larger datasets (e.g., ALIGN, JFT-300M).

\begin{table}[hbt]
\label{t3}
\begin{center}
\caption{Performance comparison of ATTIQA using various ratios of datasets.}
\resizebox{0.6\columnwidth}{!}
{
    \begin{tabular}{c||cc|cc|cc}
    \toprule
    \multirow{2}{*}{Methods} & \multicolumn{2}{c}{CLIVE~\cite{ghadiyaram2015massive}} & \multicolumn{2}{c}{KonIQ~\cite{hosu2020koniq}} & \multicolumn{2}{c}{SPAQ~\cite{fang2020perceptual}}\\
    & SROCC & PLCC& SROCC & PLCC& SROCC & PLCC \\
    \midrule
    20\% & 0.887 & 0.904 & 0.935 & 0.946 & 0.923 & 0.928 \\
    50\% & 0.889 & 0.913 & 0.938 & 0.949 & 0.924 & 0.928 \\
    \midrule
    100\% & \textbf{0.898} & \textbf{0.916} & \textbf{0.942} & \textbf{0.952} & \textbf{0.926} & \textbf{0.930} \\
    \bottomrule
    \end{tabular}
}
\label{table:dataset_scale}
\vspace{-3mm}
\end{center}

\end{table}

\begin{table}[t]
\label{t3}
\begin{center}
\caption{performance comparison of fine-tuning which only utilizes ATTIQA's each attribute head.}
\resizebox{0.6\columnwidth}{!}
{
    \begin{tabular}{c||cc|cc|cc}
    \toprule
    \multirow{2}{*}{Methods} & \multicolumn{2}{c}{CLIVE} & \multicolumn{2}{c}{KonIQ} & \multicolumn{2}{c}{SPAQ}\\
    & SROCC & PLCC& SROCC & PLCC& SROCC & PLCC \\
    \midrule
    Sharpness & 0.890 & 0.910 & 0.939 & 0.950 & 0.925 & 0.929 \\
    Contrast & 0.870 & 0.905 & 0.937 & 0.947 & 0.923 & 0.928 \\
    Brightness & 0.880 & 0.905 & 0.936 & 0.948 & 0.923 & 0.926 \\
    Colorfulness & 0.881 & 0.906 & 0.934 & 0.945 & 0.924 & 0.927 \\
    Noisiness &  0.878 & 0.901 & 0.939 & 0.950 & 0.924 & 0.928\\
    \midrule
    ATTIQA & \textbf{0.898} & \textbf{0.916} & \textbf{0.942} & \textbf{0.952} & \textbf{0.926} & \textbf{0.930} \\
    \bottomrule
    \end{tabular}
\label{table:attribute_head}
}
\end{center}
\end{table}
\subsection{Impact of attribute head}
To evaluate the impact of each attribute head, we conduct experiments that only utilize the head's feature space. 
To implement this experiment, we only adopt a single attribute head for fine-tuning instead of concatenating the five attributes feature map. 
As shown in \Tref{table:attribute_head}, although each attribute head shows similar performance since they share a backbone, we observe a slightly better performance in the sharpness attribute head. 
In contrast, the performance of other attribute heads varies depending on the dataset. 
It can be seen that this result follows an analysis of SPAQ~\cite{fang2020perceptual}, which demonstrates that sharpness is the most highly related attribute to image quality.

\begin{figure*}[htb!]
\centering
\includegraphics[clip, trim={1.4cm 0 0 0}, width=1\textwidth]{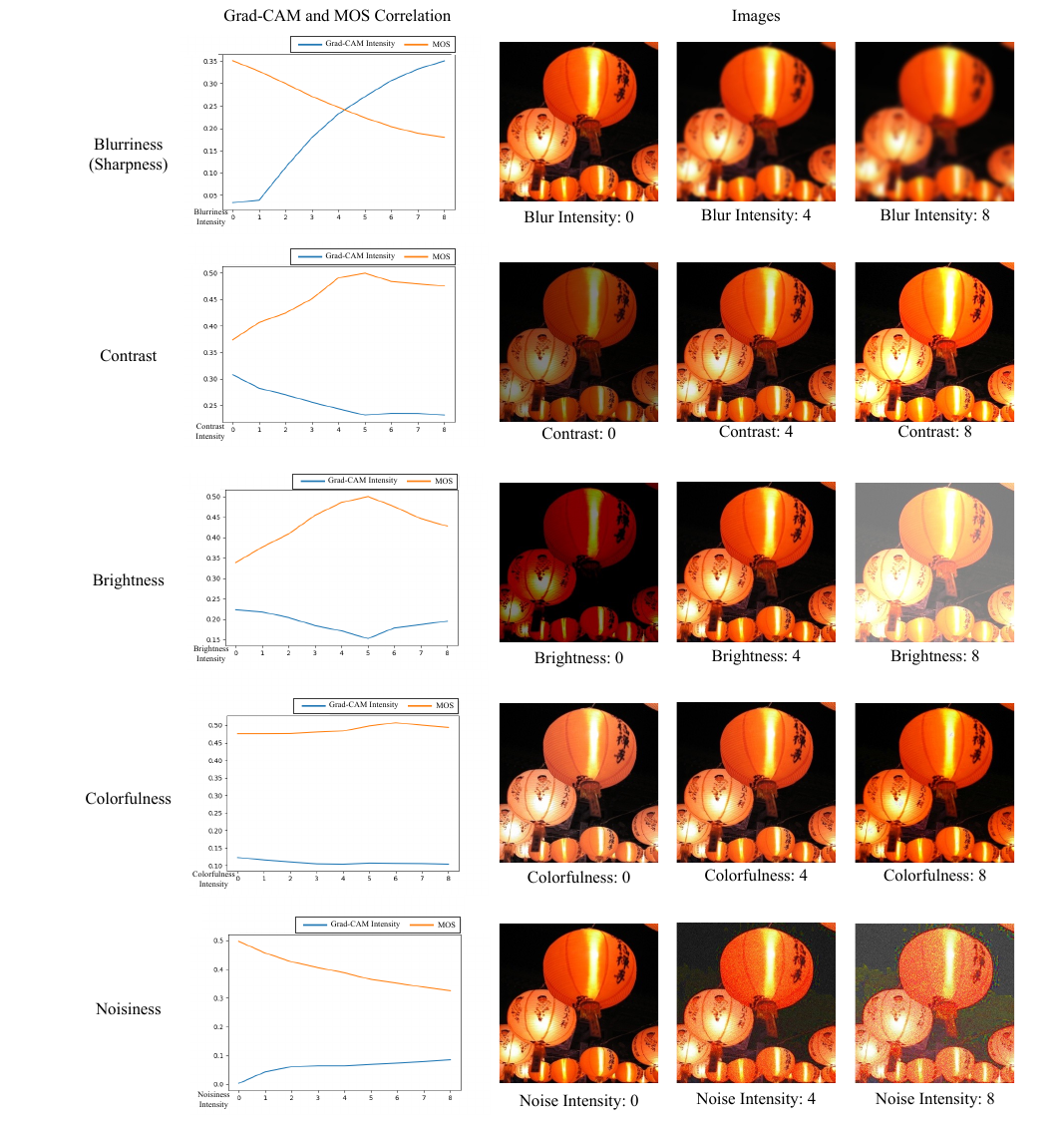}
\caption{Correlation graph illustrating the relationship between MOS and Grad-CAM across varying image attributes. On the right side, examples are provided to showcase how image attributes have been manipulated.}
\label{fig:attribute_analysis}
\end{figure*}

Moreover, as we denoted at Sec \textcolor{red}{4.5}, we conduct an additional experiment to verify that each attribute head accurately captures its corresponding attributes.
To provide a detailed explanation, we manipulate each attribute by sequentially attaching relevant distortions, described in Sec \ref{sec:prompt_selection}, to a given image and assess the significance of features by measuring the Grad-CAM from the MOS prediction to each feature map. 
Furthermore, to preserve the integrity of each feature map, we conduct this experiment using a linear probing setup. 
As shown in \Fref{fig:attribute_analysis}, we first analyze the tendency of MOS variation in response to changes in image attribute. 
For sharpness and noisiness, we can observe that MOS decreases as more noise and blur are applied to the image.
For contrast and brightness, the highest MOS is achieved when the attributes are at an optimal medium level, with the MOS decreasing when the attributes become either too high or too low.
A similar result is observed for colorfulness, but the pattern of overall variation exhibits a slight difference.
Interestingly, a negative correlation between the Mean Opinion Score (MOS) and Grad-CAM is observed for every attribute.  
It suggests that each attribute has a more significant impact on MOS prediction when a particular attribute varies to the extent that compromises the image quality.
This observation indicates that each attribute head of ATTIQA can effectively capture changes in specific attributes from the original image.

\begin{table}[htb!]
\label{t3}
\begin{center}
\caption{Performance comparison with LIQE.}
\resizebox{0.8\columnwidth}{!}
{
    \begin{tabular}{c|c||c|c|c|c|c|c||c|c|c}
    \toprule
    &\multirow{2}{*}{Methods} & \multicolumn{6}{c}{Seen Dataset} & \multicolumn{3}{c}{Unseen Dataset} \\
    &&LIVE & CSIQ & KADID & BID & LIVEC & KonIQ & TID & SPAQ & PIPAL\\
    \midrule
    SR& LIQE & 0.970 & \textbf{0.936} & \textbf{0.930} & 0.875 & \textbf{0.904} & 0.919 & \textbf{0.811} & 0.881 & 0.478 \\
    CC&ATTIQA & \textbf{0.974} & \textbf{0.936} & 0.923 & \textbf{0.891} & 0.901 & \textbf{0.920} & 0.796 & \textbf{0.891} & \textbf{0.521}\\
    \midrule
    PL&LIQE & 0.951 & 0.939 & \textbf{0.931} & 0.900 & 0.910 & 0.908 & - & - & - \\
    CC&ATTIQA & \textbf{0.977} & \textbf{0.950} & 0.927 & \textbf{0.918} & \textbf{0.917} & \textbf{0.932} & - & - & -\\
    \bottomrule
    \end{tabular}
}
\label{table:liqe}
\end{center}
\end{table}

\subsection{Comparison with LIQE}
In the main experiments, we only carried out single-dataset based experiments with LIQE to ensure a fair comparison with other baselines.
To further assess the efficacy of our model in multiple-dataset environments, we measure ATTIQA's performances utilizing a dataset tailored for LIQE and its training recipe.
As shown in Table \ref{table:liqe}, though ATTIQA is trained only using MOS, unlike LIQE which utilizes additional prompt-based annotations, it exhibits overall superior performances on given datasets.
Exceptions are observed with two datasets, TID and KADID, which primarily focus on specific distortions. 
These cases benefit from LIQE's approach of utilizing an additional dataset that contains additional annotations on synthetic distortions.

\subsection{Experiments on Synthetic Distortion-based Dataset}
To further validate the robustness of ATTIQA, we propose an additional experiment with synthetic distortion-based datasets.
We utilize LIVE, CSIQ, TID2013, and KADID-10k datasets, which measure MOS based on synthetic distortions.
As shown in \Tref{table:synthetic}, ATTIQA exhibits the best or second-best performance across all datasets.
On the KADID-10k dataset, we note that our method shows comparable results to CONTRIQUE, which uses the KADID-700k dataset at the pretraining stage.
This experiment supports the robustness of ATTIQA, demonstrating that that its representation contains information about both synthetic and authentic distortions.

\begin{table}[!tb]
\label{t3}
\begin{center}
\caption{Performances on Synthetic Distortion IQA Dataset}
\resizebox{0.8\columnwidth}{!}
{
    \begin{tabular}{c||cc|cc|cc|cc}
    \toprule
    \multirow{2}{*}{Methods} & \multicolumn{2}{c}{LIVE} & \multicolumn{2}{c}{CSIQ} & \multicolumn{2}{c}{TID2013} & \multicolumn{2}{c}{KADID-10k}\\\
    & SROCC & PLCC& SROCC & PLCC& SROCC & PLCC & SROCC & PLCC\\
    \midrule
    CONTRIQUE & 0.962 & 0.964 & 0.945 & \textbf{0.955} & \underline{0.885} & \underline{0.883} & \textbf{0.919} & \textbf{0.919} \\
    Re-IQA & 0.966 & 0.968 & \underline{0.946} & \textbf{0.955} & 0.877 & \underline{0.883} & 0.912 & 0.910 \\
    CLIP & \underline{0.968} & \underline{0.970} & 0.944 & 0.934 & 0.858 & 0.858 & 0.898 & 0.897 \\
    CLIP-IQA+ & 0.919 & 0.922 & 0.886 & 0.899 & 0.856 & 0.862 & 0.792 & 0.790 \\
    \midrule
    ATTIQA & \textbf{0.970} & \textbf{0.972} & \textbf{0.948} & \underline{0.953} & \textbf{0.898} & \textbf{0.903} & \underline{0.917} & \underline{0.917} \\
    \bottomrule
    \end{tabular}
}
\label{table:synthetic}
\end{center}

\end{table}

\section{Details on Application}
\subsection{Metrics for Generative Model}
\begin{figure*}[t]
\centering
\includegraphics[clip, trim={0 0 0 0}, width=0.9 \textwidth]{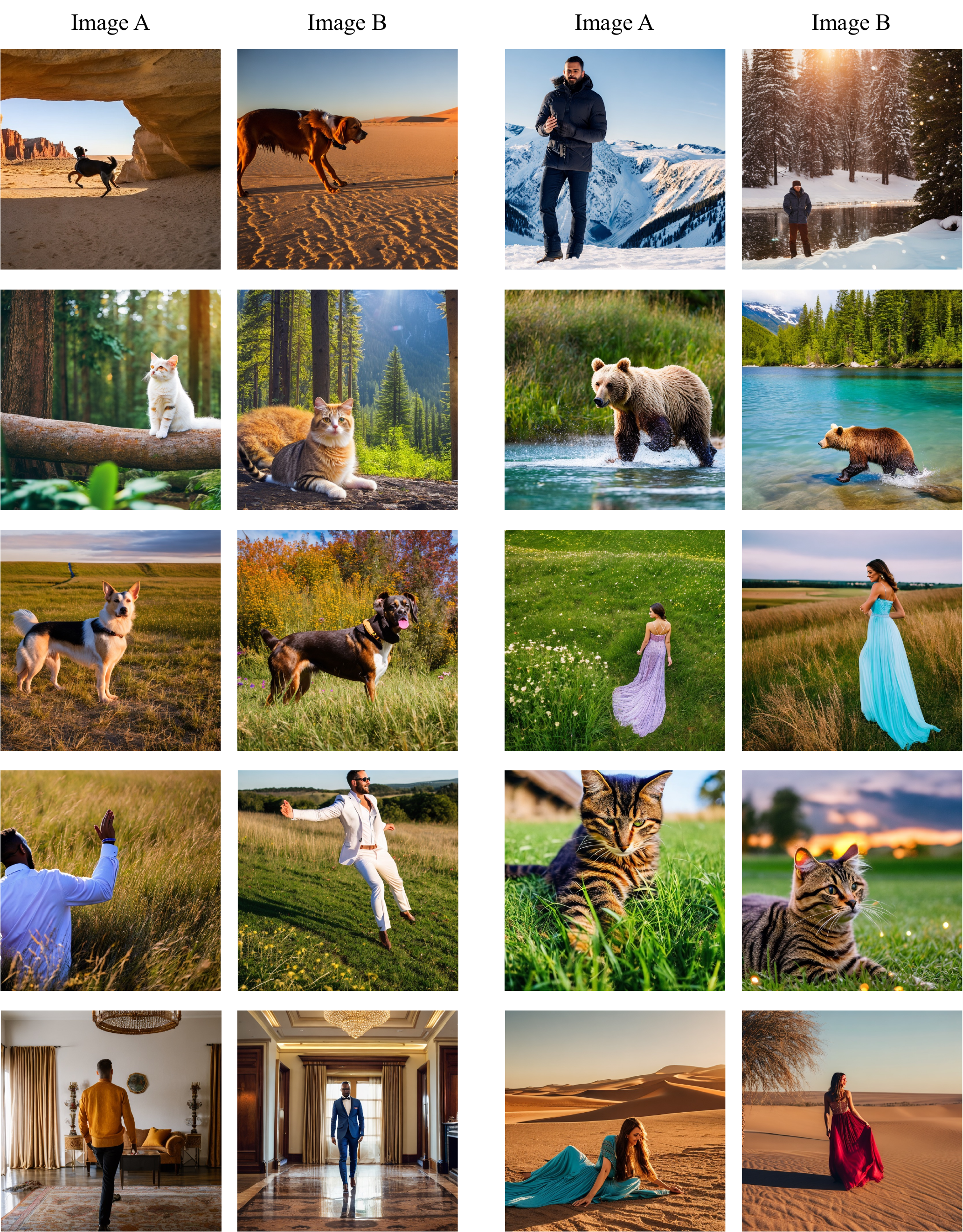}
\caption{Examples of generated images. Each paired image is synthesized using the same prompt. Image A is the image that CONTRIQUE and Re-IQA predict the high quality score. In contrast, Image B is preferred by humans and assigned a high MOS by ATTIQA.}
\label{fig:supple_diffusion}
\end{figure*}

This section provides details on the image generation pipeline used in a user study and more examples of results. For the generative model backbone, we utilize Stable Diffusion-2.1~\cite{rombach2022high}, a widely used text-to-image synthesis model. To create diverse images, we establish prompt template as \textit{``masterpiece, best quality, photorealistic photography, crystal clear, 8K UHD, \{action\} \{object\} \{place\}"} and generate images by replacing \textit{"action"}, \textit{"object"}, and \textit{"place"} respectively with various candidates reported in \Tref{table:diffusion_prompt}. \Fref{fig:supple_diffusion} shows additional qualitative comparison results of a user study. Note that this user study aims to evaluate the quality of generated images solely by showing only the two images to the subject without any information on the corresponding prompts.

\subsection{Details on User Study}
We conduct a user study using Amazon Technical Turk(AMT), gathering subjects in an anonymous setting without bias about gender or nationality.
For the survey, the descriptions were presented as follows:

\begin{itemize}
    \item[] 1) Participate in the image quality preference voting for tuning images. You can vote for the image you prefer between the two provided images
    \item[] 2) choose a better one with good image quality(color, sharpness, expressiveness..)
\end{itemize}

\subsection{Image Enhancement}
\begin{figure*}[htb!]
\centering
\includegraphics[clip, trim={0 0 0 0}, width=0.9 \textwidth]{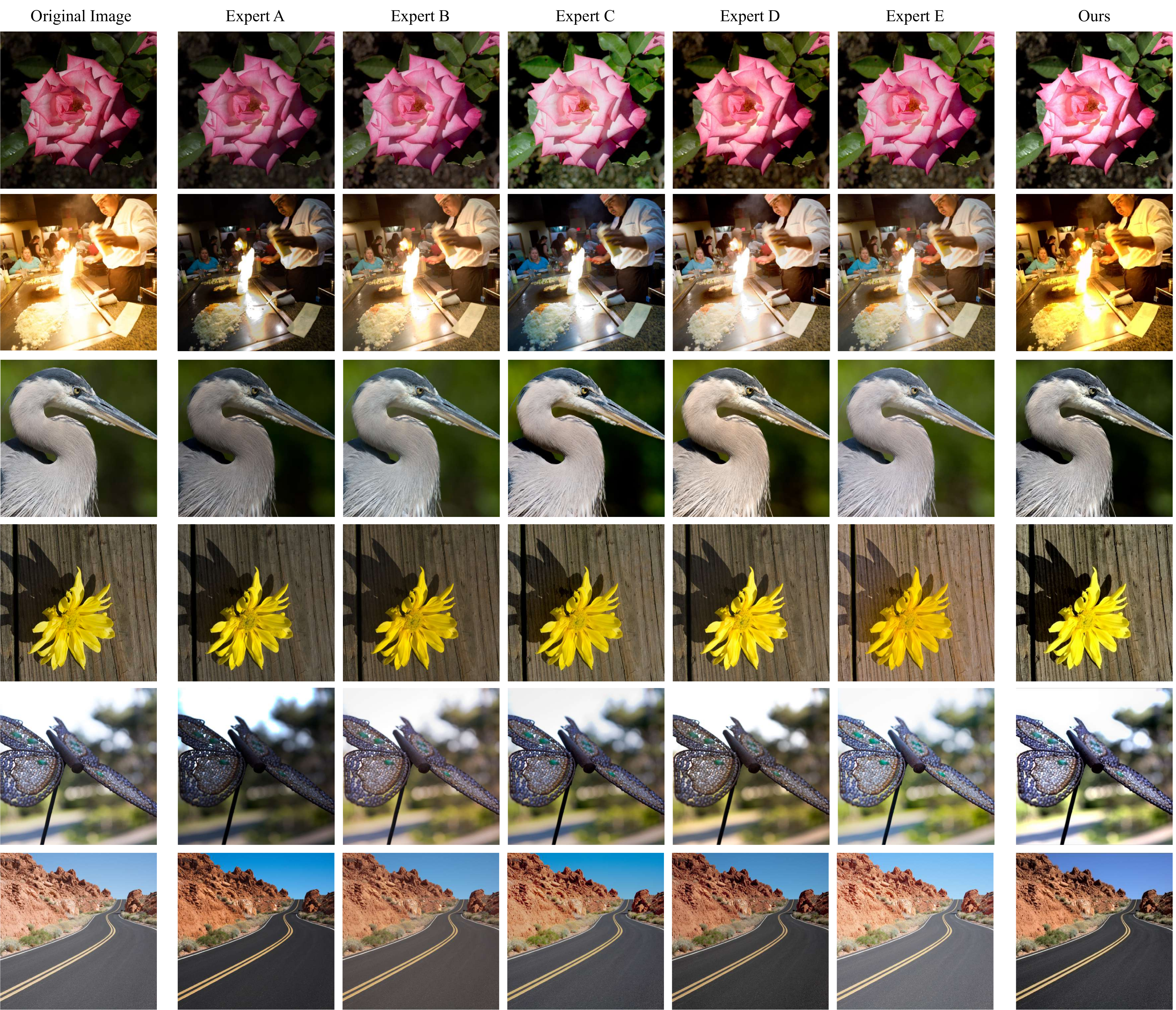}
\caption{More qualitative comparisons about image enhancement.}
\label{fig:supple_enhancement}
\end{figure*}

In this section, we explain another application in ATTIQA's MOS that is used as a reward for reinforcement learning to find optimal parameters for the ISP \cite{shin2022drl}. The ISP pipeline consists of 14 modules, of which 16 tuning parameters of 7 modules were auto-tuned from the perspective of maximizing the ATTIQA's MOS. The tuned seven modules are as follows, and the numbers in parentheses of each module represent the number of tuning parameters: Tone-mapping (3), UV channel Denoising (3), Y channel Denoising (2), Brightness/Contrast control (2), Hue/Saturation control (2), Sharpening (3), and Texture enhancing (1). The agent of the reinforcement learning was trained using 730 raw images that were 4000x3000-sized. In the user survey, we uniformly sampled 200 raw images among 5000 raw images for fair comparisons. \Fref{fig:supple_enhancement} shows additional qualitative comparison results. It can also be seen that our pipeline retouches images to make them more colorful and vivid compared to both retouching by expert C and the default settings.
\begin{table}[t]
\label{t3}
\begin{center}
\caption{Prompt candidates utilized in image generation.}
\resizebox{0.6\columnwidth}{!}
{
    \begin{tabular}{c||c}
    \toprule
    Target & Prompt candidates \\
    \midrule
    Action & \makecell{\textit{``playing", ``running", ``sleeping", ``eating", ``walking",} \\ \textit{``standing", ``sitting", ``jumping", ``dancing"}}\\
    \midrule
    Object  & \textit{``a dog", ``a cat", ``a clothed man", ``a dressed woman", ``a bear"}\\
    \midrule
    Place & \makecell{\textit{``in the grass", ``in the room", ``in the forest",} \\ \textit{``in the water", ``in the snow", ``in the desert"}}\\
    \bottomrule
    \end{tabular}

}
\label{table:diffusion_prompt}
\end{center}
\end{table}

\begin{table*}[t]
\caption{Prompt candidates for each image attribute.}
\begin{center}
\resizebox{\linewidth}{!}
{
    \begin{tabular}{c||c|c}
    \toprule
    Attribute & Prompt type&Prompt candidates \\
    \midrule
    \multirow{2}{*}{\makecell{\\ \\ \\ Sharpness}}&positive&
    \makecell{\textit{``crisp",``sharp",``defined",``clear",``distinct",``vivid",``bright",``detailed",``refined",``pristine",}\\\textit{``flawless",``lucid",``exact",``polished",``pure",``radiant",``sleek",``smooth",``resolute",``immaculate",}\\
      \textit{``brilliant",``vibrant",``rich",``clean",``meticulous",``unblemished",``sublime",``superior",``splendid",``exquisite",}\\
      \textit{``true-to-life",``tactile",``textured",``illuminated",``lustrous",``glossy",``granular",``pinpoint",``spot-on",``focused",}\\
      \textit{``unambiguous",``concise",``intense",``high-definition",``lifelike",``bold",``harmonious",``stunning",``undistorted"}}\\
      \cmidrule{2-3}
      & negative & \makecell{
      \textit{ "blurry",``fuzzy",``hazy",``vague",``indistinct",``muddled",``obscured",``smudged",``cloudy",``dull",} \\
      \textit{``muted",``out-of-focus",``pixelated",``jagged",``noisy",``grainy",``mottled",``muddy",``murky",``dim",} \\
      \textit{``foggy",``shadowy",``bleary",``washed-out",``weak",``gloomy",``clouded",``patchy",``shrouded",``veiled",} \\
      \textit{``lacking clarity",``rough",``distorted",``tarnished",``ill-defined",``ambiguous",``flat",``listless",``pale",``insipid",}\\
      \textit{``smeared",``streaked",``stained",``splotchy",``spotty",``blotchy",``dingy",``drab",``tainted"}} \\
    \midrule
    \multirow{2}{*}{\makecell{\\ \\ \\ Contrast}}&positive &\makecell{
      \textit{``crisp",``defined",``vivid",``sharp",``clear",``distinguished",``bold",``pronounced",``high-contrast",``distinct",} \\
      \textit{``lucid",``striking",``intense",``robust",``dynamic",``stark",``rich",``deep",``emphasized",``highlighted",} \\
      \textit{``vibrant",``solid",``notable",``prominent",``enhanced",``powerful",``contrastive",``standout",``bright",``conspicuous",} \\
      \textit{``discernible",``marked",``potent",``compelling",``dramatic",``forceful",``illuminated",``brilliant",``radiant",``meticulous",} \\
      \textit{``articulate",``impressive",``splendid",``magnified",``amplified",``accentuated",``divergent",``outstanding",``captivating"} \\
    }\\
    \cmidrule{2-3}
    & negative & \makecell{
      \textit{``muddy",``flat",``blurred",``faded",``indistinct",``washed-out",``lackluster",``dull",``weak",``subdued",} \\
      \textit{``undistinguished",``low-contrast",``ambiguous",``pale",``muted",``obscure",``vague",``insipid",``clouded",``nebulous",} \\
      \textit{``tenuous",``confusing",``feeble",``diminished",``indiscernible",``unnoticeable",``slight",``unemphasized",``shadowed",``doubtful",} \\
      \textit{``hazy",``unsaturated",``ill-defined",``unremarkable",``bleak",``insignificant",``bland",``monotone",``uniform",``muddled",} \\
      \textit{``equivocal",``unaccentuated",``listless",``understated",``unimpressive",``nondescript",``faint",``impotent",``inaudible",``discreet"} \\
    } \\
    \midrule
    \multirow{2}{*}{\makecell{\\ \\ \\ Brightness}}&positive  & \makecell{
      \textit{``luminous",``bright",``vivid",``brilliant",``radiant",``gleaming",``illuminated",``clear",``sparkling",``shiny",} \\ 
      \textit{``glowing",``light-filled",``dazzling",``lucent",``resplendent",``shimmering",``lustrous",``beaming",``crisp",``vibrant",} \\
      \textit{``intense",``well-lit",``brillante",``glittering",``glistening",``blazing",``effulgent",``reflective",``aglow",``incandescent",} \\
      \textit{``high-key",``fiery",``lambent",``twinkling",``opulent",``sunlit",``burnished",``pristine",``flashing",``undimmed",} \\
      \textit{``sunny",``spotlit",``blinding",``flawless",``translucent",``glossy",``crystal-clear",``immaculate",``gleamy"} \\
    }\\
    \cmidrule{2-3}
    & negative & \makecell{
      \textit{``dim",``dull",``dark",``shadowy",``obscured",``faint",``gloomy",``pale",``muted",``clouded",} \\
      \textit{``bleak",``underexposed",``drab",``faded",``murky",``shaded",``veiled",``flat",``dusky",``tenebrous",} \\
      \textit{``sombre",``gray",``lackluster",``washed-out",``overcast",``smoky",``subdued",``muffled",``eclipsed",``sullen",} \\
      \textit{``unlit",``opaque",``low-key",``blurred",``darkened",``blackened",``shadowed",``misty",``lightless",``moonless",} \\
      \textit{``starless",``inky",``twilight",``foggy",``overclouded",``cimmerian",``umbrous",``pitch-dark"} \\
    }\\
    \midrule
    \multirow{2}{*}{\makecell{\\ \\ \\ Colorfulness}}& positive & \makecell{
      \textit{``vibrant",``rich",``vivid",``saturated",``brilliant",``lively",``radiant",``bold",``bright",``colorful",} \\
      \textit{``intense",``resplendent",``lush",``deep",``dazzling",``varied",``dynamic",``electric",``illuminated",``vibrantly-hued",``multicolored",} \\
      \textit{``kaleidoscopic",``strong",``fluorescent",``fiery",``prismatic",``stunning",``flashy",``beaming",``pigmented",``chromatic",} \\
      \textit{``glistening",``spectacular",``polychromatic",``sunny",``iridescent",``opulent",``rainbow-like",``effulgent",``color-laden",``invigorating",} \\
      \textit{``gorgeous",``lustrous",``gleaming",``dramatic",``bursting",``captivating",``energetic"} \\
    }\\
    \cmidrule{2-3}
    & negative & \makecell{
      \textit{``drab",``dull",``washed-out",``muted",``faded",``pale",``flat",``monochrome",``grayish",``uninspiring",} \\
      \textit{``lifeless",``bleak",``tarnished",``insipid",``blurred",``cloudy",``dim",``neutral",``colorless",``lackluster",} \\
      \textit{``subdued",``murky",``dusty",``dusky",``undistinguished",``muddy",``unsaturated",``shadowed",``overcast",``veiled",} \\
      \textit{``bland",``indistinct",``unvaried",``uniform",``faint",``anaemic",``vague",``wan",``stale",``ashen",} \\
      \textit{``pastel",``watered-down",``sallow",``obscured",``indeterminate",``discolored",``ill-defined",``tinged",``hazy"} \\
    }\\
    \midrule
    \multirow{2}{*}{\makecell{\\ \\ \\ Noisiness}}&positive  & \makecell{
     \textit{``clear",``crisp",``smooth",``pure",``sharp",``pristine",``flawless",``noiseless",``unblemished",``intact",} \\
      \textit{``clean",``polished",``immaculate",``well-defined",``impeccable",``spotless",``untainted",``sleek",``neat",``unspoiled",} \\
      \textit{``intact",``refined",``clutter-free",``lucid",``undisturbed",``unmarred",``untarnished",``perfect",``refreshing",``distinct",} \\
      \textit{``vivid",``bright",``detailed",``accurate",``faithful",``exquisite",``superb",``top-notch",``first-rate",``matchless",} \\
      \textit{``peerless",``masterful",``skillful",``uncompromised",``optimal",``optimum",``superior",``prime",``finest"} \\
    }\\
    \cmidrule{2-3}
    & negative &  \makecell{
      \textit{``grainy",``speckled",``mottled",``patchy",``dirty",``blemished",``marred",``flecked",``spotty",``noisy",} \\
      \textit{``blurry",``fuzzy",``hazy",``cloudy",``splotchy",``streaky",``dotted",``gauzy",``scratchy",``scattered",} \\
      \textit{``dappled",``distorted",``messy",``smudged",``lackluster",``gloomy",``dim",``overcast",``pockmarked",``crackled",} \\
      \textit{``choppy",``erratic",``spattered",``discolored",``inconsistent",``irregular",``shoddy",``subpar",``mediocre",``unrefined",} \\
      \textit{``vague",``ambiguous",``dull",``dreary",``stained",``blemish-ridden",``defective",``imperfect",``second-rate",``tarnished",``degraded"} \\
    }\\
    \bottomrule
    \end{tabular}
}
\label{table:prompt_candidates}
\end{center}
\end{table*}

\clearpage

%
%
\bibliographystyle{splncs04}
\bibliography{egbib}